# The Pedestrian Patterns Dataset

**Kasra Mokhtari and Alan R. Wagner**

Mechanical Engineering Department and Aerospace Engineering Department,
The Pennsylvania State University
Email: kbm5402@psu.edu, azw78@psu.edu

**Abstract**

We present the pedestrian patterns dataset for autonomous driving. The dataset was collected by repeatedly traversing the same three routes for one week starting at different specific timeslots. The purpose of the dataset is to capture the patterns of social and pedestrian behavior along the traversed routes at different times and to eventually use this information to make predictions about the risk associated with autonomously traveling along different routes. This dataset contains the Full HD videos and GPS data for each traversal. Fast R-CNN pedestrian detection method is applied to the captured videos to count the number of pedestrians at each video frame in order to assess the density of pedestrians along a route. By providing this large-scale dataset to researchers, we hope to accelerate autonomous driving research not only to estimate the risk, both to the public and to the autonomous vehicle but also accelerate research on long-term vision-based localization of mobile robots and autonomous vehicles of the future.

## Introduction

Autonomous driving started in the 1980s, when Carnegie Mellon University (CMU, Pittsburgh, PA) presented its Navlab vehicles that operated in structured environments (Thorpe, Herbert et al. 1991). In recent years, autonomous cars and the prospect that they will populate our roads in the near future has attracted increasing attention. These vehicles promise a number of benefits to society, including reducing the number of road accidents due to human error, reducing traffic congestion and increasing convenience (Anderson, Nidhi et al. 2014).

Real-world data plays a crucial role in the development, testing and validation of autonomous vehicles algorithms prior to their deployment on public roads. A number of vision-based autonomous driving datasets have been released (Brostow, Fauqueur et al. 2009, Dollár, Wojek et al. 2009, Pandey, McBride et al. 2011, Geiger, Lenz et al. 2013, Pfeiffer, Gehrig et al. 2013, Blanco-Claraco, Moreno-Dueñas et al. 2014). These datasets are primarily used to develop algorithms for autonomous vehicles focusing on particular aspects of autonomous driving such as motion estimation (Nistér, Naroditsky et al. 2006, Geiger, Ziegler et al. 2011), pedestrian and vehicle detection (Viola, Jones et al. 2005, Benenson, Omran et al. 2014), semantic classification (Posner, Cummins et al. 2008, Long, Shelhamer et al. 2015), and localization in the same environment under different conditions (McManus, Upcroft et al. 2015, Linegar, Churchill et al. 2016). The dataset we present is unique in that it was recorded along three different specific routes chosen for because their differences in pedestrian density. Moreover, our dataset was recorded at specific times of the day that temporal changes in pedestrian density over times of the day and days of the week could be compared. To the best of our knowledge, there is no publicly available dataset has been collected by repeatedly traversing exactly the same route at different times of a day and different days of the week.

In this paper, we present a dataset for autonomous driving that traverses three different routes over the course of a week at the same time each day. The data consists of more than 600GB of videos and GPS data recorded by Samsung Galaxy S9 Plus that was mounted to the windshield of a car. The data was collected by repeatedly traversing three different routes between two locations in State College, Pennsylvania as shown in Figure 1 over a period of three weeks (one week for each route and six times a day) in November 2018. By driving three different routes under different conditions and different times of a day, we capture a large range of variation in scene appearance and structure due to illumination, weather, congestion, and the placement and movement of objects. By providing this-large-scale dataset to researchers, we hope to accelerate autonomous driving research not only to estimate the risk, both to the public and to the autonomous vehicle but also accelerate research on long-term vision-based localization of the mobile robots and autonomous vehicles of the future.

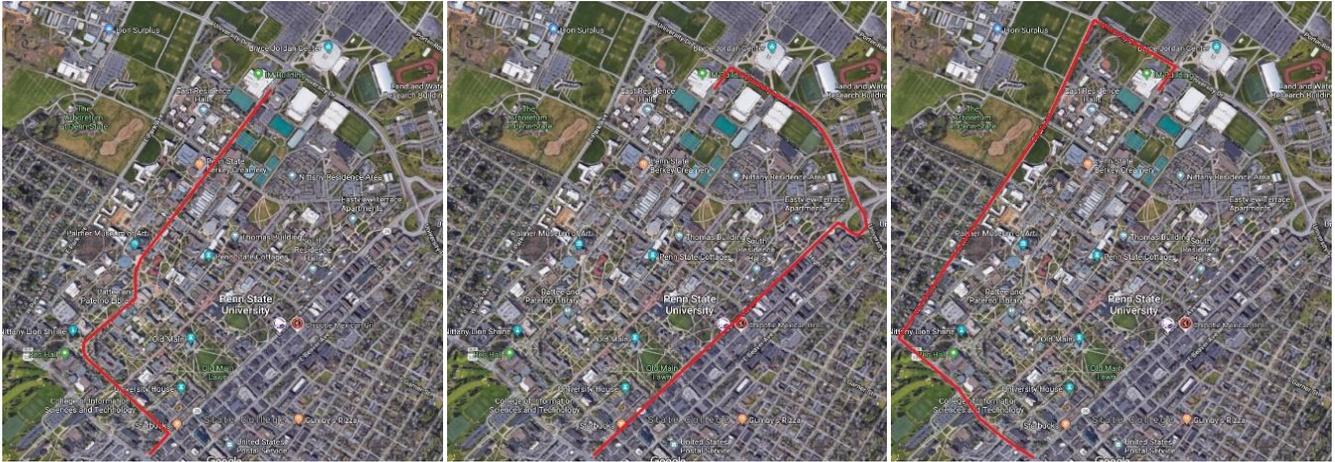

Figure 1: Three routes between the same location in State College, PA. Route A (left), route B (middle) and route C (right). The car traversed each route for 42 times including 7 days and six timeslots a day as: 8:45, 10:45, 12:45, 14:45, 16:45 and 17:45.

## Data Collection

Data was collected over the period of three weeks in November 2018 and consists of over 500 miles of recorded manually driving a 2010 Honda Accord in State College, PA. The vehicle was driven at an approximate average speed of 20 mph along three predetermined routes (Figure 1). These routes were selected because they offered variations in terms of urban versus rural driving. One week of driving was devoted to collecting data along each different route. Data was collected over the course of seven days at six different times. The following six times were selected for data collection: 8:45, 10:45, 12:45, 14:45, 16:45 and 17:45. The total dataset therefore consists of 126 different traversals.

The dataset consists of videos and GPS locations collected by mounting a Samsung Galaxy S9 Plus to the windshield of the car. This device has two back cameras; firstly, as a "wide angle camera" and secondly, as a "telephoto camera". The wide-angle camera is a super speed dual pixel 12MP AF sensor with sensor size as 1/2.55", pixel size 1.4 μm and sensor ration 4:3. However, telephoto camera is 12MP AF sensor with sensor size 1/3.6", pixel size 1.0 μm and sensor ration 4:3. The device automatically switched between these two cameras depending on the lighting conditions to optimize the video generated. Captured videos are Full HD ($1920p \times 1080p$) and 60 frame-per-seconds. To reduce the buildup of dust and moisture on the camera lens, they were cleaned with a microfiber cloth before each traversal. The total uncompressed size of the raw data is 600GB. GPS data was collected using "GPSLogger" application on Android with 1Hz frequency.

The data for each traversal is labeled either as sunny, cloudy, rainy for easy organization by weather conditions. Figure 2. presents the different condition labels and the number of traversals with each label. Figure3. shows the distribution of videos length for route A, route B and route C, respectively. Figure 4. presents a series of images taken from the same location on different traversals, denoting the range of appearances changing over time. A MATLAB development tool for accessing and manipulating data is also available for download along with the raw data at: https://sites.psu.edu/real/PedestrianPatternsDataset.

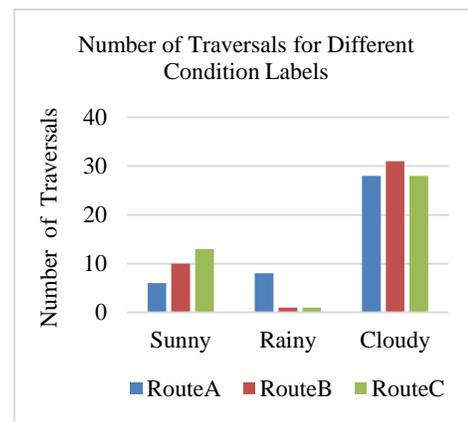

Figure 2: Number of traversal for different condition including sunny, rainy and cloudy weather. Traversals are labeled by their weather condition.

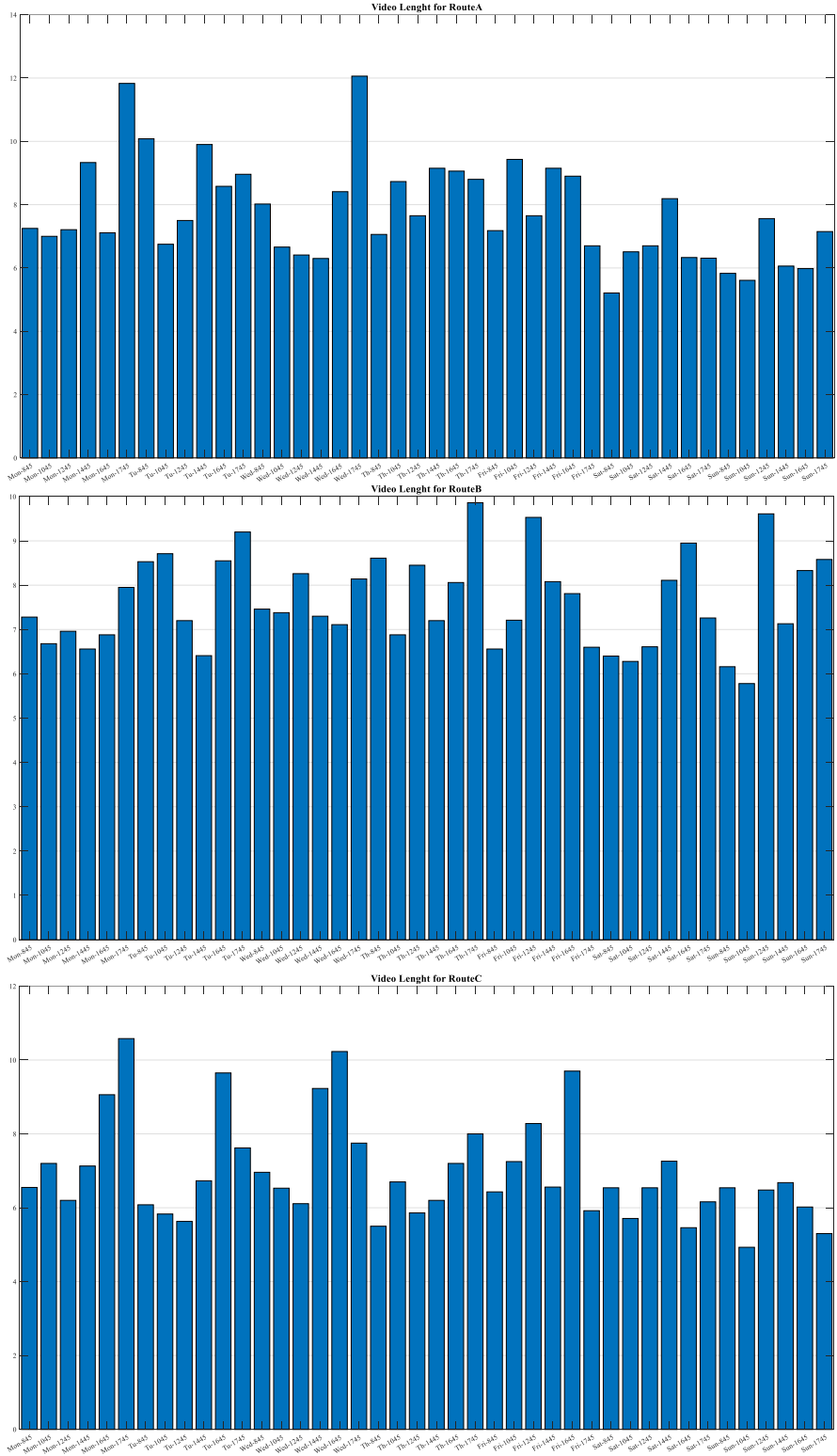

Figure 3: Distribution of Videos length (min) for Route A, Route B and Route C

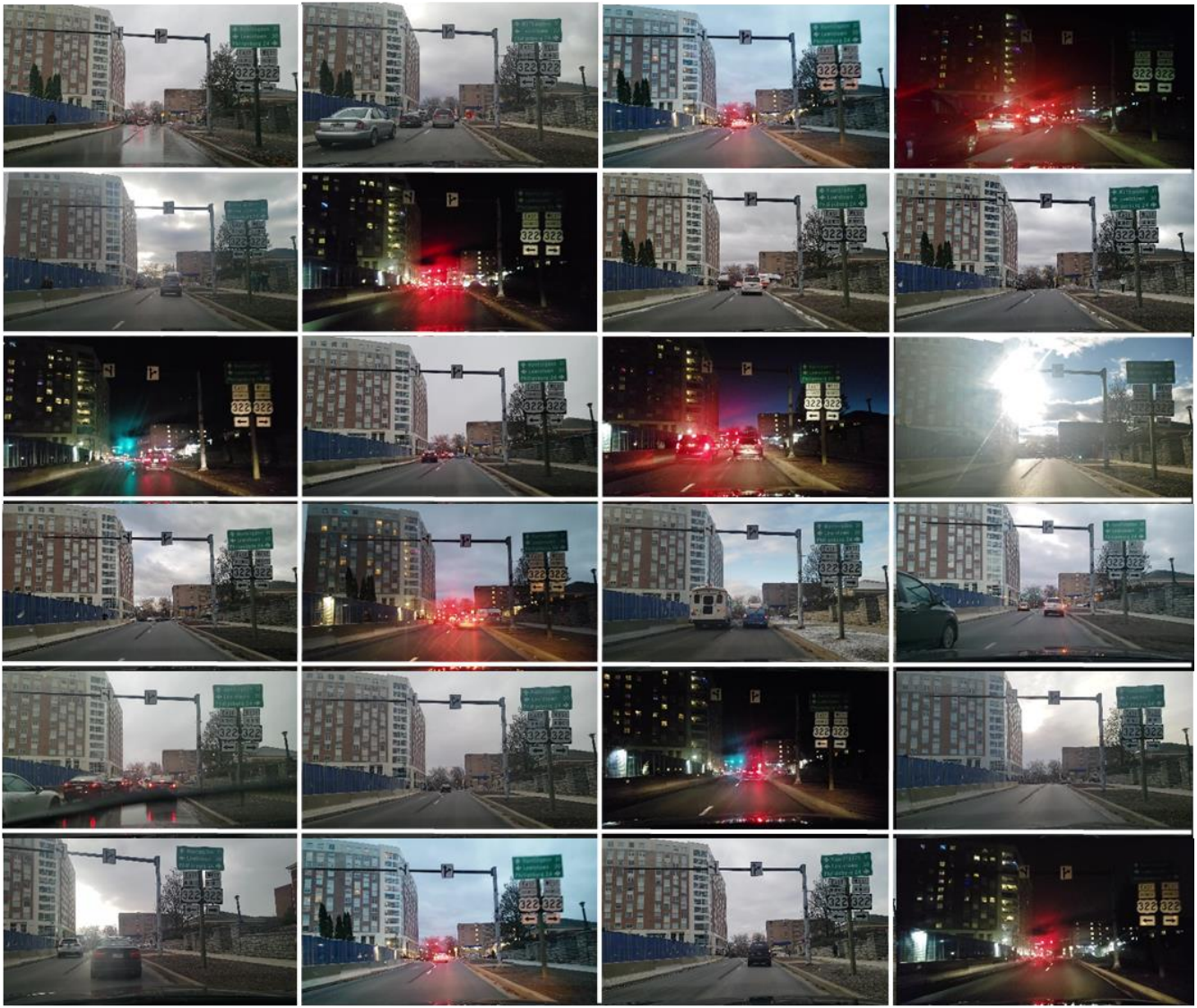

Figure 4: Series of images of the same location denoting the large changes in appearances including lighting and weather Condition

## Characterizing the Social Environment

Autonomous vehicles may need to interact with pedestrians, police officers, traffic workers, bicyclists, and others. We hypothesize that prior knowledge, both of the number and type of people, along a route can be used by an autonomous vehicle to select the least risky path. One purpose of this dataset is to provide information about the probability of encountering people while driving along different routes and to develop algorithms that will allow an autonomous vehicle to characterize the types of people that the vehicle may encounter during a trip and their behavior. As discussed below, our goal is to use this dataset to estimate the risk, both to the public and to the autonomous vehicle, associated with traveling along a particular route. For example, an autonomous vehicle might use this data to predict the number of young children that it will encounter along a route and use this information to select a different route in order to reduce the possibility of harming a child if an accident were to occur.

In order to characterize the social environment, the number of pedestrians at each location was first counted using the Fast R-CNN for pedestrian detection (Girshick 2015). Pedestrian detection is used to generate bounding boxes any pedestrians found in an image as shown in Figure 5. Fast R-CNN is a fast and accurate object detection algorithm. The implementation of Fast R-CNN code used in this paper was written in Python (Felzenszwalb, Girshick et al. 2010) and is available as open-source at https://github.com/rbgirshick/fast-rcnn. Figure 6 depicts a heat map of the pedestrians detected along each different route.

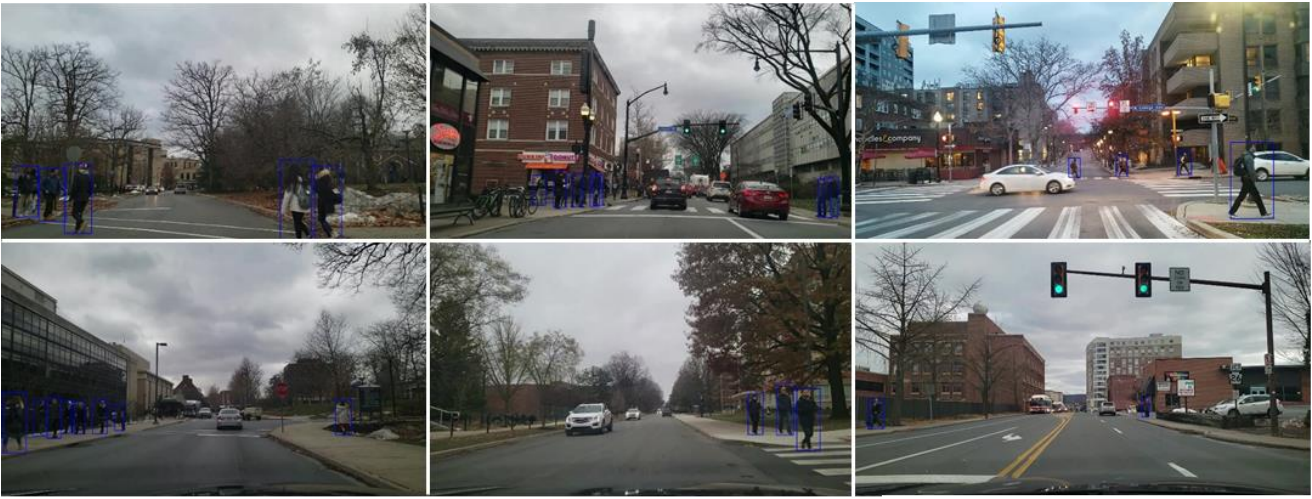

Figure 5: Blue bounding boxes around the pedestrian instances using Fast RCNN pedestrian detection method

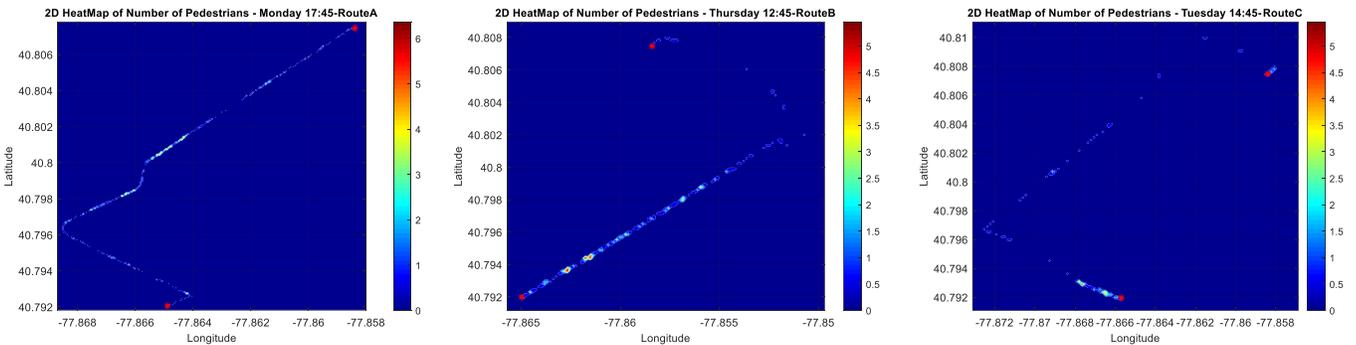

Figure 6: 2D heat map of pedestrians for Monday 17:45 route A (left), Thursday 12:45 route B (middle) and Tuesday 8:45 route C (right)

The Fast-RCNN python code was run on Ubuntu 18.04.02 LTS using NVIDIA GeForce GTX 1070. Most of the videos are 7-10 minutes long and required approximately 20 minutes to process. Data for detected pedestrian at each frame are then combined with GPS data resulting into a 2D heat map of pedestrians for each traversal.

2D heat maps of pedestrians are shown in Figure 6. for the Monday 17:45 route A, Thursday 12:45 route B and Tuesday 8:45 route C, respectively. Table 1 present the total number of frames and total number of frames with detected pedestrians for route A, route B and route C, respectively. Captured videos for route A and route B have approximately 1.15M frames. The videos for route A have the higher number of frames with pedestrians because this route mainly goes through Penn State campus.

## Risk Analysis for Understanding When to Trust

Trust is closely related to risk (Engle-Warnick and Slonim 2006). Mayer notes that vulnerability is a facet of trust (Mayer, Davis et al. 1995). For trust in an autonomous system, Wagner defines trust with respect to risk stating, "Trust is defined as a belief, held by the trustor, that the trustee will act in a manner that mitigates the trustor's risk in a situation in which the trustor has put its outcomes at risk"(Wagner 2009) . Risk can be defined in terms of the expected loss, or cost, associated with choosing an action $x$.

|  | Route A | Route B | Route C |
|---|---|---|---|
|  | ~ 1.15M (~595K) | ~ 1.15M (~340K) | ~ 1.05M (~120K) |
| Total frames in Days (with pedestrian) | ~ 740K (~410K) | ~ 760K (~220K) | ~ 660K (~85K) |
| Total frame in nights (with pedestrian) | ~ 275K (~185K) | ~ 255K (~120K) | ~ 545K (~35K) |

Table 1: Total number of frames and total number of frames with pedestrians for route A, route B and route C.

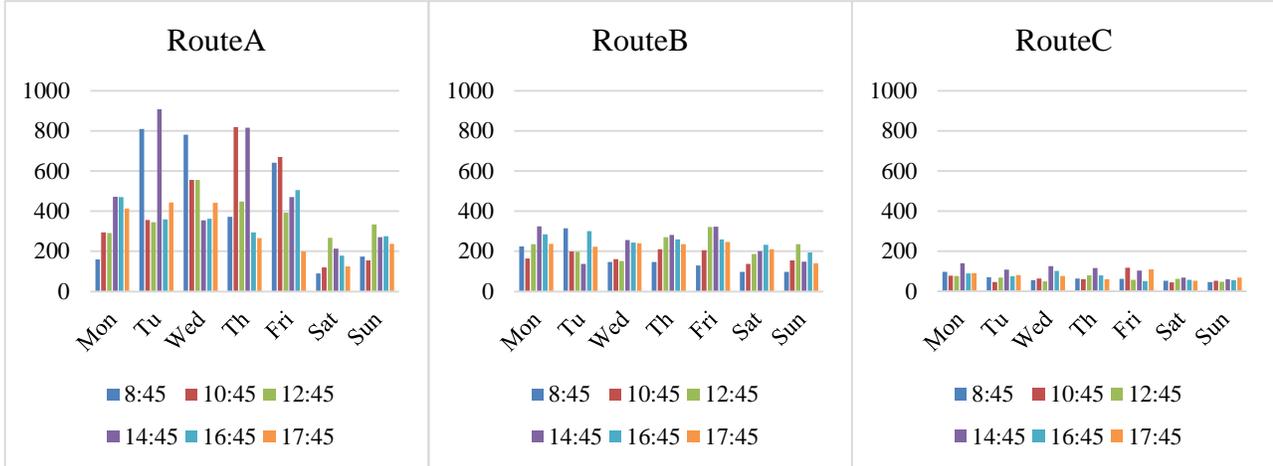

Figure 7: Total number of pedestrians for each traversal.

$$R(x) = \sum_{y} L(x,y)p(y) \quad (1)$$

where $L(x, y)$ the loss associated with choosing action $x$ given event $y$ occurs, and $p(y)$ is the probability of event $y$ occurring. The event $y$ corresponds to various possible failures of an autonomous system.

The total number of pedestrian for each traversal can be determined by integrating the number of pedestrians over the route. Total number of pedestrians for each traversal is shown in Figure 7. For example, if one wants to travel from the starting point to the ending point on at 17:45 on Friday, the optimal route will be route C since it has the least number of pedestrians reflecting the least amount of risk. As another example, if one wants to traverse along route B on Friday, traveling at 8:45 a vehicle will encounter the least total number of pedestrians across the six existing timeslots.

## Data Organization

For distribution we have divided up the datasets into individual routes. Three route folders (route A, route B and route C) have seven subfolders corresponding to the seven days of the week and each day has six subfolders denoting the six timeslots. Within the timeslots folder the format for each data type are as follows:

- Raw video: These raw videos are $1920p \times 1080p$ and 60 frame-per-seconds. To reduce download file sizes, we have further divided each traversal video into pieces approximately 10 minutes in length. The files are structured as <Route>/<day>/<Timeslot>.mp4.

- Processed videos: This is the output video generated running the original video through the pedestrian detection python code. Blue bounding boxes are put around the detected pedestrians. These processed videos are $1920p \times 1080p$ and 20 frame-per-seconds. The files are structured as <Route>/<day>/<Timeslot>-python.mp4.

- GPS file: This file contains the GPS data for each traversal and is structured as: <Route>/<day>/<Timeslot>.gpx.

- Excel file: This file describes the total number of detected pedestrians in each frame and the pixel number of the bounding box vertices for each detected pedestrian.

## Summary and Future Work

This paper has presented a new dataset capturing patterns of pedestrian presence and behavior along different driving routes. The dataset is meant to serve as a tool for the research community interested in understanding and predicting patterns of pedestrian behavior in order to use this information to increase the safety of future autonomous vehicles.

Our own future work will offer tools and algorithms that utilize this dataset for the purpose of estimating risk along particular routes during different times of the day and days of the week. For instance, a Bayesian Network model for an autonomous vehicle can be designed in order to estimate the probability of fatal accident or a fatal injury due to weather condition, time of a day, age or sex of the drivers, etc. The risk analysis mentioned in this paper then, can be integrated with this Bayesian Network model of the autonomous vehicle in order to have the better understanding of the autonomous vehicle risk assessment counting on different factors including road conditions, driver's characteristics and pedestrians by which the vehicle is passing. We also intend to develop methods that use behavior and age recognition to evaluate if the presence of specific types of people (such as children) or specific types of behavior (jaywalking) increases the risk of an accident and should be avoided. Finally, we believe that patterns in density of pedestrians can be used as a tool to help the vehicle determine its location. To this end, we will apply Monte Carlo Localization and Neural Network methods for path identification only by implementing the pedestrian pattern data.

## Acknowledgments

This work was supported by Air Force Office of Scientific Research contract FA9550-17-1-0017.

## References

Anderson, J. M., K. Nidhi, K. D. Stanley, P. Sorensen, C. Samaras and O. A. Oluwatola (2014). *Autonomous vehicle technology: A guide for policymakers, Rand Corporation.*

Benenson, R., M. Omran, J. Hosang and B. Schiele (2014). Ten years of pedestrian detection, what have we learned? *European Conference on Computer Vision, Springer.*

Blanco-Claraco, J.-L., F.-Á. Moreno-Dueñas and J. González-Jiménez (2014). "The Málaga urban dataset: High-rate stereo and LiDAR in a realistic urban scenario." *The International Journal of Robotics Research* **33**(2): 207-214.

Brostow, G. J., J. Fauqueur and R. Cipolla (2009). "Semantic object classes in video: A high-definition ground truth database." *Pattern Recognition Letters* **30**(2): 88-97.

Dollár, P., C. Wojek, B. Schiele and P. Perona (2009). "Pedestrian detection: A benchmark."

Engle-Warnick, J. and R. L. Slonim (2006). "Learning to trust in indefinitely repeated games." Games and Economic Behavior **54**(1): 95-114.

Felzenszwalb, P. F., R. B. Girshick, D. McAllester and D. Ramanan (2010). "Object detection with discriminatively trained part-based models." *IEEE transactions on pattern analysis and machine intelligence* **32**(9): 1627-1645.

Geiger, A., P. Lenz, C. Stiller and R. Urtasun (2013). "Vision meets robotics: The KITTI dataset." *The International Journal of Robotics Research* **32**(11): 1231-1237.

Geiger, A., J. Ziegler and C. Stiller (2011). Stereoscan: Dense 3d reconstruction in real-time. *2011 IEEE Intelligent Vehicles Symposium (IV), Ieee.*

Girshick, R. (2015). Fast r-cnn. *Proceedings of the IEEE international conference on computer vision.*

Linegar, C., W. Churchill and P. Newman (2016). Made to measure: Bespoke landmarks for 24-hour, all-weather localisation with a camera. *2016 IEEE International Conference on Robotics and Automation (ICRA), IEEE.*

Long, J., E. Shelhamer and T. Darrell (2015). Fully convolutional networks for semantic segmentation. *Proceedings of the IEEE conference on computer vision and pattern recognition.*

Mayer, R. C., J. H. Davis and F. D. Schoorman (1995). "An integrative model of organizational trust." *Academy of management review* **20**(3): 709-734.

McManus, C., B. Upcroft and P. Newman (2015). "Learning place-dependant features for long-term vision-based localisation." *Autonomous Robots* **39**(3): 363-387.

Nistér, D., O. Naroditsky and J. Bergen (2006). "Visual odometry for ground vehicle applications." *Journal of Field Robotics* **23**(1): 3-20.

Pandey, G., J. R. McBride and R. M. Eustice (2011). "Ford campus vision and lidar data set." *The International Journal of Robotics Research* **30**(13): 1543-1552.

Pfeiffer, D., S. Gehrig and N. Schneider (2013). Exploiting the power of stereo confidences. *Proceedings of the IEEE Conference on Computer Vision and Pattern Recognition.*

Posner, I., M. Cummins and P. Newman (2008). "Fast probabilistic labeling of city maps." *Proceedings of Robotics: Science and Systems IV, Zurich, Switzerland.*

Thorpe, C., M. Herbert, T. Kanade and S. Shafer (1991). "Toward autonomous driving: the cmu navlab. i. perception." *IEEE expert* **6**(4): 31-42.

Viola, P., M. J. Jones and D. Snow (2005). "Detecting pedestrians using patterns of motion and appearance." *International Journal of Computer Vision* **63**(2): 153-161.

Wagner, A. R. (2009). The role of trust and relationships in human-robot social interaction, Georgia Institute of Technology.